# Improving Negative Sampling for Word Representation using Self-embedded Features

Long Chen
University of Glagow
Glasgow, UK
long.chen@glasgow.ac.uk

Fajie Yuan*
University of Glagow
Glasgow, UK
f.yuan.1@research.gla.ac.uk

Joemon M Jose
University of Glagow
Glasgow, UK
joemon.jose@glasgow.ac.uk

Weinan Zhang
Shanghai Jiao Tong University
Shang Hai, China
wnzhang@sjtu.edu.cn

## ABSTRACT

Although the word-popularity based negative sampler has shown superb performance in the skip-gram model, the theoretical motivation behind oversampling popular (non-observed) words as negative samples is still not well understood. In this paper, we start from an investigation of the gradient vanishing issue in the skip-gram model without a proper negative sampler. By performing an insightful analysis from the stochastic gradient descent (SGD) learning perspective, we demonstrate that, both theoretically and intuitively, negative samples with larger inner product scores are more informative than those with lower scores for the SGD learner in terms of both convergence rate and accuracy. Understanding this, we propose an alternative sampling algorithm that dynamically selects informative negative samples during each SGD update. More importantly, the proposed sampler accounts for multi-dimensional self-embedded features during the sampling process, which essentially makes it more effective than the original popularity-based (one-dimensional) sampler. Empirical experiments further verify our observations, and show that our fine-grained samplers gain significant improvement over the existing ones without increasing computational complexity.

## 1 INTRODUCTION

In recent years, there has been a surge of work proposed to represent words as dense vectors, using various training methods inspired from neural-network language modeling [3, 5, 40]. These representations, referred to as "neural embedding" or "word embedding", have been shown to perform well in a variety of natural language processing (NLP) tasks, such as named entity recognition [15, 31], sentiment analysis [28, 37] and question answering [48].

One of the most popular word embedding techniques is the skip-gram model. Given a corpus of target words and their context, it aims to predict the probability of observing a context word conditioned on a target word by sliding a symmetric window over a subsampled training corpus. One of the major difficulties of these language models is that one needs to compute activation functions by summing over an entire vocabulary, which is often millions of words in scale. To reduce the computational cost, researchers often use two lines of methods, one is hierarchical softmax [27], another is noise contrastive estimation (or alternatively, negative sampling) [11]. While useful in general, the effectiveness of such methods largely depends on the assumption that oversampling frequent words would lead to better performance since they are more informative than less frequent ones [11]. However, in fact, infrequent words may also carry important information. In addition, a simple global and static sampling method such as popularity-based sampling strategy cannot effectively handle the cases where words are represented by a large number of embedded features.

To tackle the aforementioned problems, we first show that a not well-designed (e.g. random) sampler would easily result in the gradient vanishing problem during the parameter learning process, especially when the corpus size is very large and the words are long tail distributed. Hence, most SGD updates have no effect, which leads to slow convergence for the learning algorithm. Both theoretical and experimental analysis reveals that popularity-based negative sampling is able to alleviate the vanishing gradient issue. However, our analysis also shows that popularity-based negative sampling can only achieve suboptimal performance for two reasons: (1) non-observed context words with high popularity (frequency) are often irrelevant to the target word; (2) popular words are sampled without considering the dynamic change of parameters in the training process. Hence, in this paper we propose a non-popularity sampling strategy, termed as *Adaptive Sampler*, which makes useful of multi-dimensional semantic and syntactic information, and samples top ranked context words by considering both embedding variables and the current state of SGD learner. On two real-world corpora, the proposed algorithm can significantly outperform the

*The first two authors contributed equally to this work and share the first authorship.

original word2vec baseline. Furthermore, our method has an amortized constant runtime without increasing time complexity of the original word2vec [23].

The rest of this paper is organized as follows. We firstly introduce the related work in Section 2. Section 3 formally defines the problem of word embedding with adaptive sampling. Section 4 systematically presents the proposed word-embedding sampler. The experimental results and analysis are reported in Section 5. Finally, we present our conclusion and future work in Section 6.

## 2 RELATED WORK

Neural network language models [22, 24, 43] have attracted a lot of attention recently given their dense and learnable representation form and generalization property, as a contrast to the traditional bag-of-words representations. Word2vec skip-gram [23] (cf. Section 3) is arguably the most widely used word embedding models today. However, the computation of output vector (softmax layer) represents the probability of the context word and is the size of the entire vocabulary [6], which is computationally prohibitive (even with the recent advance of GPU-accelerated computing). This has been a thorny problem ever since Bengio's seminal work of neural network language model [3].

There are several ways to tackle this challenge. A common way is hierarchical softmax, which was first proposed by Mnih and Hinton [25], where a hierarchical tree is constructed to index all the words in a corpus as leaves for the prediction of the normalized probability of the target class [27]. Peng recently proposed an incremental training method which is able to learn the softmax tree faster than global training [30] while the performance of this model is still comparable to the original version.

Another popular way to reduce the computational cost is simply selecting only a small fraction of the output's dimensions, which are either randomly or heuristically chosen. The reconstruction sampling of Dauphin et al. [8], the efficient use of biased importance sampling in [20], the adoption of noise contrastive estimation [11] in Mnih and Kavukcuoglu [26] all belong to this category. The most famous one in this line of work is arguable negative sampling (cf. Section 3), which is the simple version of noise contrastive estimation (NCE) that randomly samples the words not in the context to distinguish the observed data from the artificially generated noise. Empirically, negative sampling generally outperforms hierarchical softmax, especially for frequent words [23]. The reason is that hierarchical softmax builds a tree over the whole vocabulary, and the leaf nodes representing rare words will inevitably inherit their parent vector representations in the tree, as a result, they are affected by other frequent words in the corpus. Thus, we choose negative sampling as baseline in this work due to its superior performance.

Recently, the use of approximate maximum-inner-product Function has become popular [8, 41] to select a good candidate subset, which is somewhat similar to our idea. But our approach upgraded the inner product function into a rank-invariant function, and thus is computationally more efficient than these alternatives. In addition, they use the function for the task of image recognition [44] and recommender systems [33, 46, 47], while the feasibility and effectiveness of this approach for the task of word embeddings is still largely unknown.

More generally speaking, matrix factorization (MF) model is also employed to reduce the dimension of a co-occurrence matrix. Context-distribution smoothing MF [19] and global MF [31] (also known as GloVe) all belong to this category. While generally effective, MF models actually employ negative sampling implicitly [18], and thus these two techniques tend to perform quite similarly for most downstream NLP tasks [2]. To the best of our knowledge, this paper is the first attempt to investigate SGD update at a finer-grained level with embedding features, and propose to use an adaptive sampler. Furthermore, the proposed samplers can be easily adopted to other more complex factorization models, such as tensor factorization [39], even though we only implement them on word2vec in this paper.

## 3 PRELIMINARY

First we formally introduce several concepts and notations. Then we shortly recapitulate the skip-gram[1] model with negative sampling (SGNS). The novel contribution of this section is to show the theoretical motivation behind oversampling popular (non-observed) words as negative samples.

### 3.1 Continuous Skip-gram Model

In [23], words are trained with an unlabelled corpus of words $w_1, w_2, ..., w_n$ (usually $n$ is about millions) and the context for word $w_i$ are words surrounding it in a $T$-sized window $w_{i-T}, ..., w_{i-1}, w_{i+1}, ..., w_{i+T}$. The corpus of observed word and context pairs is denoted as $D$. We use $\#(w, c)$ to denote the frequency of pair $(w, c)$ appears in $D$.

Each target word $w$ corresponds to a vector $\vec{w} \in \mathbb{R}^d$ and similarly each context word $c$ is represented as a vector $\vec{c} \in \mathbb{R}^d$, where $d$ is the embedding dimension. The values in the embedding vector referred to as latent variables are the parameters to be learned. The vector $\vec{w}$ is the row in a $|V| \times d$ matrix $W$, and vectors $\vec{c}$ is a row in a $|V| \times d$ matrix $C$, where $|V|$ is the vocabulary size and is derived from the corpus $D$. In such cases, $W_i$ and $C_i$ represent vector representations of the $i$-th target word and context word in the vocabulary respectively.

Our starting point is the skip-gram embedding model trained with the negative sampling. Consider a word-context pair $(w, c)$. Let $p(D = 1|w, c)$ be the probability that $(w, c)$ is observed in $D$, and $p(D = 0|w, c) = 1 - p(D = 1|w, c)$ the probability that $(w, c)$ is non-observed. The distributions can then be expressed as:

$$p(D = 1|w, c) = \sigma(\vec{w} \cdot \vec{c}) = \frac{1}{1 + e^{-\vec{w} \cdot \vec{c}}} \tag{1}$$

where $\vec{w}$ and $\vec{c}$ are $d$ dimensional vectors, and will be learned by the model.

The word embedding learning algorithm aims to maximize $p(D = 1|w, c)$ for observed pair $(w, c)$, meanwhile, minimize $p(D = 0|w, c)$ for randomly sampled non-observed pairs, under the intuition that randomly sampled non-observed word-context pairs are more likely to be negative pairs. For each observed $(w, c)$ pair and a set of $k$ negative examples $V^-_{w_k}$ that are sampled from the whole negative example set $V^-_w$, the SGNS objective function is defined as [23]:

[1]We merely elaborate our idea by using the skip-gram model, while it simply applies to the continuous bag-of-words (CBOW) model.

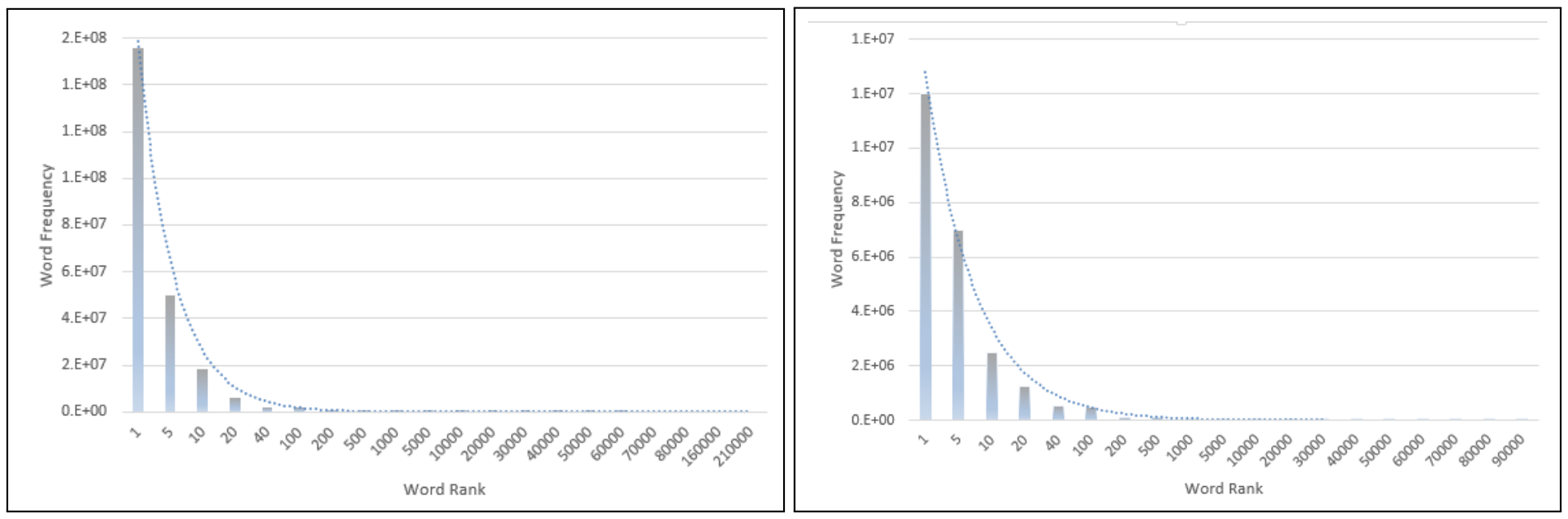

**Figure 1: (a) and (b) show the word popularity distribution of Wiki2017 and NewsIR datasets respectively. Popularity in both datasets is tailed.**

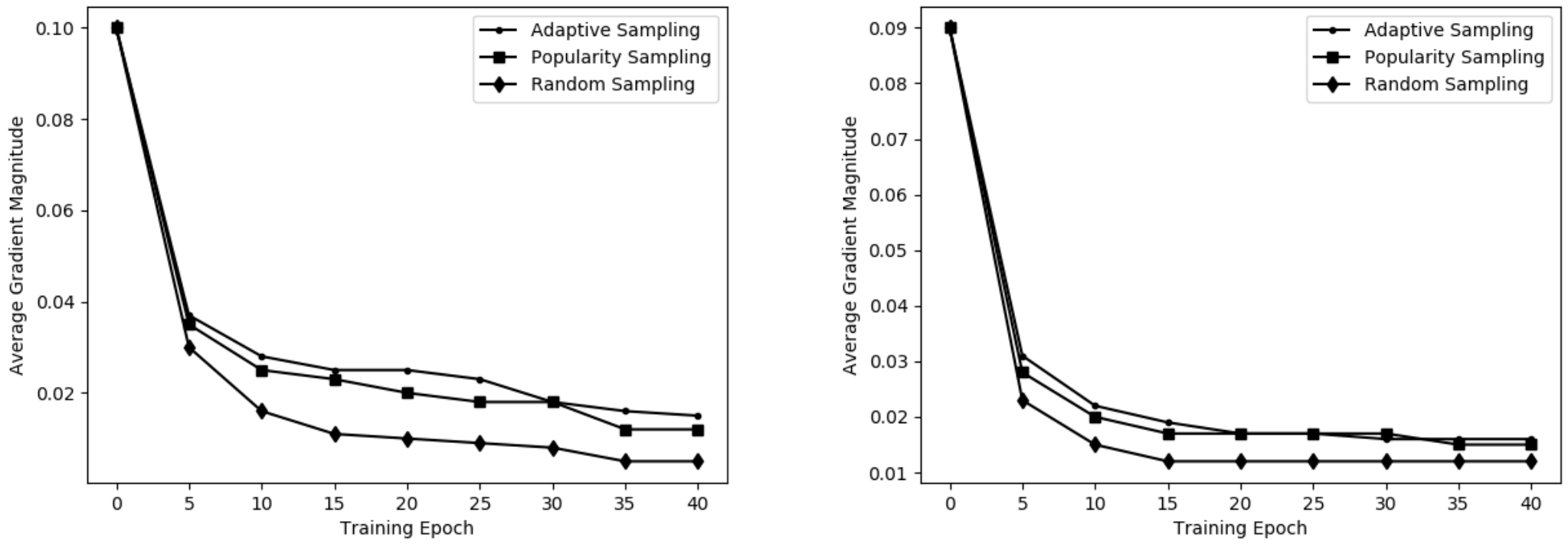

**(a)** **(b)**

**Figure 2: (a) and (b) show the probability of gradient magnitude of varying samplers with skip-gram model on Wiki2017 and NewsIR datasets respectively.**

$$E = -\log \sigma(\vec{w} \cdot \vec{c}_P) - \sum_{c_N \in V^-_{w_k}} \log \sigma(-\vec{w} \cdot \vec{c}_N) \tag{2}$$

where $c_P$ is the positive (observed) context word and $c_N$ is the negative (non-observed) context word for $w$, which is selected by oversampling popular words.

$$p_D(c_N) = \frac{\#(c_N)^\alpha}{|D|} \tag{3}$$

where $\#(c_N)$ is the frequency of word $c_N$ in the corpus $D$, and $|D|$ represents the number of available words. The exponent $\alpha$ controls the weight distribution of sampled negative words, which is experimentally shown that when $\alpha = 0.75$, the algorithm performs the best [23]. Note that the random sampling distribution is a special case by setting $\alpha = 0$.

Minimizing this objective function makes observed pair $(w, c)$ have similar embedding representation while scattering the non-observed ones. This is intuitively correct as words appear in similar context should bear a close resemblance. Mathematically, SGNS tries to maximize the inner-product of similar words and minimize the dot-product of dissimilar ones.

## 3.2 Gradient Issues in Tailed Word Distribution

Even though the word-popularity based negative sampler (i.e. Eq. 3) has been successfully applied in various word embedding models, the theoretical motivation behind oversampling popular negative words is yet known. In the following, we seek to show, both theoretically and intuitively, drawing negative words with high popularity frequency is a reasonable yet suboptimal sampling method.

To begin with, we follow Mikolov et al. [23] by using stochastic gradient descent (SGD) to optimize Eq. 2. The gradient for model parameter $\theta$ is then given as:

$$\frac{\partial E}{\partial \theta} = (\sigma(\vec{w} \cdot \vec{c}_P) - 1)\frac{\partial \vec{w} \cdot \vec{c}_P}{\partial \theta} + \sum_{c_N \in V^-_{w_k}} \sigma(\vec{w} \cdot \vec{c}_N)\frac{\partial \vec{w} \cdot c_N}{\partial \theta} \tag{4}$$

Let $\theta = c_P$ or $\theta = c_N$, then we can update $\vec{c}_P$ and $\vec{c}_N$ with Eq. 4 as follows:

$$\vec{c}_P^{\,new} \leftarrow \vec{c}_P^{\,old} - \eta \underbrace{(\sigma(\vec{w} \cdot \vec{c}_P) - 1)}_{\triangle_{w,c_P}} \vec{w} \tag{5}$$

$$\vec{c}_N^{\,new} \leftarrow \vec{c}_N^{\,old} - \eta \underbrace{\sigma(\vec{w} \cdot \vec{c}_N)}_{\triangle_{w,c_N}} \vec{w} \tag{6}$$

where $\triangle_{w,c_P}$ and $\triangle_{w,c_N}$ are known as gradient magnitude. Note that since the number of observed words (i.e. $c_P$) is very small compared with non-observed words (i.e. $c_N$), it does not require to design a special sampling method. Hence, in this paper we focus only on the learning process of Eq. 6.

To provide the insight and motivation of popularity based negative sampling, we analyze the gradient update process of Eq. 6 by employing a simple random sampler (i.e. $\alpha = 0$ in Eq. 3). First, we observe that the value of the updated gradient in Eq. 6 is largely dependent on the the score function (i.e. $\vec{w} \cdot \vec{c}_N$)). The quantity of $\triangle_{w,c_N}$ is obviously a probability and is close to 0 if $c_N$ is correctly predicted as a true negative word, because in this case $\vec{w} \cdot \vec{c}_N$ is supposed to be small. In fact, the gradient magnitude $\triangle_{w,c_N}$ can be understood as how much influence the $(w, c_N)$ pair has for improving $\Theta$. If it is close to 0, nothing is learned from the pair $(w, c_N)$ because its gradient vanishes, i.e. $\theta$ cannot be changed in the updating process. It is worth noting that $\triangle_{w,c_N}$ relies on parameters $\Theta$ and is constantly changed during learning. Hence, we proceed by analyzing how the vanishing gradient occurs without a proper negative sampler and why oversampling popular words is able to address the issue.

In word-embedding tasks, word popularity (i.e. occurrence frequency) is typically non-uniform distributed and some words are in general more popular than others. Figure 1 shows word frequency distributions for Wiki2017 and NewsIR datasets respectively. Both datasets shows that the vast majority of words are low popularity and thus by random sampling, most selected negative words are those tailed words. On the other side, $\triangle_{w,c_N}$ is supposed to be small in general if $c_N$ has lower popularity (i.e. a lower rank position) in Figure 1. The reason is straightforward because an ideal learner is expected to assign a larger score for $\triangle_{w,c}$ if $c$ is an observed positive context word and a lower score if $c$ is a negative word. As we know, the lower popularity $c$ has, the fewer times it acts as a positive context word, and thus the lower score $\triangle_{w,c}$ is assigned to. If $\triangle_{w,c}$ has a very small value, then $\sigma(\vec{w} \cdot \vec{c_P})$ is close to 0, which means the gradient vanishes. Hence, the purpose of oversampling popular words is to select more informative negative examples to overcome gradient vanishing problem and speed up the training process. Figure 2 shows the gradient magnitude of varying sampling approaches. It can be seen that after a few training epochs, almost all the negative samples, selected by the random sampler, have very small gradient magnitudes ($\triangle_{w,c_N}$), which suggests most of them are useless in the SGD learning process. On the contrary, the popularity-based sampler and our proposed adaptive sampler (cf. Section 4) can significantly increase the gradient magnitude by a large factor, and thus can alleviate the gradient vanishing issue.

## 3.3 Learning Optimal Ranking for Embeddings

In this subsection, we provide an intuitive example to explain the merits of popularity oversampling from ranking perspective. The reason is that training word embedding can also be naturally viewed as a ranking task that ranks an observed context word $c_P$ higher than any non-observed context word $c_N$ [14]. To illustrate this, we give a schematic of a ranked list for a target word $w$ as below, where +1 and -1 denote an observed and non-observed context word respectively. We use NDCG (Normalized Discount Cumulative Gain [21]) as the ranking metric for explanation, similar to other metrics, e.g. AP (Average Precision) [21].

$$\text{Rank Order}: \overbrace{-1,\ -1,\ +1,\ -1,\quad \underbrace{-1,\ -1,\ +1}_{\triangle NDCG(w)_{75}=0.033}}^{\triangle NDCG(w)_{71}=0.409}\ , -1\,, \ldots, -1$$

where $\triangle NDCG_{ij}$ denotes the size of NDCG change for word $w$ when positive context word with the position $i$ and negative context word with the position $j$ get swapped. As can be seen, the value of $\triangle$ $NDCG(w)_{71}$ is much larger than that of $\triangle NDCG(w)_{75}$. This implies that $\triangle$ $NDCG(w)$ is likely to be larger if the non-observed word $c_N$ has a smaller rank. Hence, the new NDCG value after swapping is also larger if $\triangle NDCG(w)$ is larger[2]. This is intuitively correct as the high ranked non-observed words hurt the ranking performance more than the low ranked ones. A higher NDCG value for the rank list of target word $w$ corresponds with better accuracy in distinguishing observed and non-observed contexts. As discussed in Section 3.2, popular words are more likely to have larger scores (or smaller rank) than non-popular words. Our idea here is similar to that used in [46, 47] for a different problem.

In fact, both Section 3.2 and 3.3 show that larger score (smaller rank) negative context words are more informative for training the embedding models, and popular words are alternative instances for larger score negative words. Empirical results in word2vec[23] have already proven that approximate sampling based on word popularity distribution usually results in both promising accuracy and faster convergence.

## 3.4 Issues of Popularity Sampling

Both the theoretical and intuitive motivations regarding the negative sampler have been discussed: select for a target word $w$, and one (or several) negative context word $c$ such that the pair $(w, c)$ is informative at the current state of learning. However, the original popularity oversampling does not reflect this for two reasons: (1) It is static and thus the empirical popularity distribution does not change during the learning process. However, the estimated score $\hat{y}(c|w) = \vec{w} \cdot \vec{c}$ (or rank $\hat{r}(c|w)$) of a context word $c$ changes during learning. E.g. $c$ might have a larger score (with $w$) in the beginning but after several epochs of training it is ranked low. (2) The sampler is global and does not reflect the semantic and syntactic information regarding how informative a word is. For example, a popular word is more likely to act as a context word with a group of target words, but still can be irrelevant for another one. Meanwhile, learning with popularity-based sampler can slow down after the algorithm learns to (generally) rank positive context word above popular words, and thus can be inaccurate with ranking long tail but high scoring context words. Both points can also be observed in the gradient magnitude $\triangle_{w,c}$ , which depends on the inner product of self-embedded features $\vec{w}$ and $\vec{c}_N$ and changes during learning. In the next section, we will present a new adaptive sampler that select informative negative words based on the embedded features in $\vec{w}$ and $\vec{c}$, which are known as the low-dimensional representation of semantic and syntactic information.

[2] $NDCG(w)_{\text{new}}=NDCG(w)_{\text{old}}+\triangle NDCG(w)$

# 4 IMPROVED NEGATIVE SAMPLING

In this section, a dynamic sampler that takes account of multi-dimensional self-embedded features is proposed to replace the original popularity-based sampler.

## 4.1 Basic adaptive Sampler

As has been discussed in Section 3.4, we are able to propose a straightforward adaptive sampler which defines the sampling distribution directly based on the scoring function $\hat{y}(c|w) = \vec{w} \cdot \vec{c}$ instead of the popularity word distribution. Intuitively, when a negative word $c_N$ in a given word list is sampled, the closer $c_N$ is ranked at the the top position by $\hat{y}(c_N|w)$, the more important $c_N$ is. This has been understood from both the gradient magnitude $\triangle_{w,c_N}$ (Section 3.2) and ranking perspective (Section 3.3). For example, if $(w, c_N)$ is given, we should choose $c_N$ such that $\hat{y}(c_N|w)$ is large since it will largely increase both $\triangle_{w,c_N}$ and NDCG. In the following, instead of using the notion of a large score it is better to formalize a small predicted rank $\hat{r}(c_N|w)$, since largeness of scores is only a relative value to other words but ranks will be an absolute value. This allows us to formulate a basic dynamic sampling distribution that assigns higher sampling weight for small ranked context words.

$$p_D(c_N|w) \propto \exp\left(\frac{-\widehat{r}(c_N|w)}{\lambda}\right), \lambda = |V| \cdot \rho, \rho \in (0, 1] \tag{7}$$

where $\rho$ is the hyper-parameter that controls the shape of the exponential distribution and should be tuned according to the dataset.

**Properties:** The context word distribution (Eq. 7) depends on $\hat{r}(c_N|w)$, and has two important properties:

(1) Feature-dependent: Remind that $\hat{r}(c_N|w)$ is the rank of word $c_N$ among all words in the vocabulary using the inner product of self-embedded features $\vec{w}$ and $\vec{c_N}$ for ordering words, and thus it is feature-dependent and inherently can represent the semantic and syntactic relations between $w$ and $c_N$.

(2) Adaptive: The sampler changes while model parameters are learned because changes in parameters lead to consequently in changes in the scoring model $\hat{y}$, the ranking $\hat{r}(c_N|w)$, and hence, the sampler.

## 4.2 Efficient Sampling Algorithm

So far, we have designed a trivial adaptive & self-embedded feature based sampler. However, the additional computational cost of the proposed sampler is to score all non-observed context words of $w$ in the whole word list to obtain the rank $\hat{r}(c_N|w)$, which means before each SGD update, the rough computational complexity of $O(d|V_w^-|)$ is required[3]. As the whole training process has always millions of SGD updates, it is generally infeasible in practice. In this section, we will show how approximative sampling from Eq. 7 can be implemented efficiently in amortized time for the word embedding task.

Let the scoring model $\hat{y}$ still be the inner product of a factorized matrix.

$$\hat{y}(c|w) = \vec{w} \cdot \vec{c} = \sum_{f=1}^{d} \vec{w}_f \vec{c}_f \tag{8}$$

[3] The size of non-observed context words $|V_w^-|$ is much larger than that of observed ones $|V_w^+|$, i.e. $|V_w^-| \approx |V|$, as $|V_w^+| + |V_w^-| = |V|$.

**Algorithm 1:** Skip-gram model with adaptive and feature-dependent oversampling of negative words.

1: Random initialize the parameters $\Theta$
2: $t \leftarrow 0$; **while** $t < MaxIteration$ **do**
  **if** $t \% |V| \log |V| = 0$ **then**
    **for** $f \in \{1, ..., d\}$ **do**
      compute $\hat{r}(.|f)$
      compute $\vec{\sigma}_f$ and $\vec{\mu}_f$
    **end**
  **end**
  Draw $(w, c) \in D$ uniformly
  **for** $neg = 0; neg < k; neg++$ **do**
    Draw $r$ from $p(r) \propto \exp(-r/\lambda)$
    Draw $f$ from $p(f|w) \propto |\vec{w}_f|\vec{\sigma}_f$
    **if** $\text{sgn}(\vec{w}_f) = 1$ **then**
      $c_N = r^{-1}(r|f)$
    **end**
    **else**
      $c_N = r^{-1}(|V| - f + 1)|f)$
    **end**
    Store $c_N$ in $V_{w_k}^-$
  **end**
  Update $\theta \in \Theta$
  $t \leftarrow t + 1$
**end**

Now, a fast adaptive and feature-dependent sampling algorithm is presented which approximates the sampler from Eq 7. The idea is to formalize Eq 7 as a mixture of ranking distributions over normalized factors. The mixture probability is calculated by a normalized version of the scoring function Eq. 8

**Normalization Scheme:** First, we assume the context word factors for each dimension $f$ correspond to the normal distribution, then the standard factor $\vec{c'}_f \sim \mathcal{N}(0, 1)$ is given as

$$\vec{c'}_f = \frac{\vec{c}_f - \vec{\mu}_f}{\vec{\sigma}_f} \tag{9}$$

where $\vec{\mu}_f$ and $\vec{\sigma}_f$ is the mean value and standard deviation for each $f$. Hence, we update Eq 8 by replacing $\vec{c}_f$ in Eq 9.

$$\begin{aligned} \hat{y}(c|w) = \vec{w} \cdot \vec{c} &= \sum_{f=1}^{d} \vec{w}_f \vec{c}_f = \sum_{f=1}^{d} |\vec{w}_f| \text{sgn}(\vec{w}_f)(\vec{c'}_f \vec{\sigma}_f + \vec{\mu}_f) \\ &= \underline{\sum_{f=1}^{d} |\vec{w}_f| \text{sgn}(\vec{w}_f) \vec{\mu}_f} + \sum_{f=1}^{d} |\vec{w}_f| \text{sgn}(\vec{w}_f) \vec{c'}_f \vec{\sigma}_f \end{aligned} \tag{10}$$

where sgn denotes the sign function, and the first (underlined) term can be treated as a constant value relative to the context word $c$. In other words, if we want to obtain the rank $\hat{r}(c|w)$ of different context words, we just need to compare the rank of the second terms, say, $\hat{r}^*(c|w)$, in a linear transformation equation. Thus, we can derive a new scoring function $\hat{y}^*(c|w)$ which shares the same ranking relations with $\hat{y}(c|w)$.

$$\hat{y}^*(c|w) = \sum_{f=1}^{d} |\vec{w}_f| \text{sgn}(\vec{w}_f) \vec{c'}_f \vec{\sigma}_f \quad (11)$$

**Modeling the Mixture Distribution:** Before deriving the mixture distribution, we first revisit the physical meaning of Eq 12. As we known, each word is represented by an embedding vector, which means $\hat{y}^*(c|w)$ or $\hat{y}^*(\vec{c}|\vec{w})$ can be decomposed as a mixture distribution of all real-valued elements in the embedding vector. Understanding this, we can define the sampling distribution as follows:

$$p(c_N|w) = \sum_{f=1}^{d} p(f|w) p(c_N|w, f) \quad (12)$$

As discussed, the real-value $\vec{c'}_f$ follows the standard normal distribution, we can define $p(c_N|w, f)$ analogously to Eq. 7 by replacing the ranking function $\widehat{r^*}$.

$$p(c_N|w, f) \propto \exp(\frac{-\widehat{r^*}(c_N|w, f)}{\lambda}) \quad (13)$$

where $\widehat{r^*}(c_N|w, f)$ can be inferred from $\hat{y}^*(c_N|w, f)$. Now, two questions arise: how to perform sampling according to $p(f|w)$, and how to estimate the score of $\hat{y}^*(c_N|w, f)$? Since the value of $p(f|w)$ stands for the importance of the dimension $f$ for $\vec{w}$, we have

$$p(f|w) \propto |\vec{w}_f| \vec{\sigma}_f \quad (14)$$

which means the large value $\vec{w}_f$ and $\vec{\sigma}_f$ have, the more important $f$ is in $\vec{w}$. Thus, we have

$$\hat{y}^*(c_N|w) = \sum_{f=1}^{d} p(f|w) \text{sgn}(\vec{w}_f) \vec{c'}_{Nf} \quad (15)$$

Accordingly, this scoring function is reasonably given as

$$\hat{y}^*(c_N|w, f) = \text{sgn}(\vec{w}_f) \vec{c'}_{Nf} \quad (16)$$

Calculating Eq 16 is not straightforward. However, according to above analysis, $\hat{y}(c_N|w, f)$ is also the linear transformation of $\hat{y}^*(c_N|w, f)$, we can have a simpler function:

$$\hat{y}(c_N|w, f) = \text{sgn}(\vec{w}_f) \vec{c}_{Nf} \quad (17)$$

The relation of the scoring function $\hat{y}(c_N|w, f)$ and its rank $\hat{r}(c_N|w, f)$ is: the word on rank $r$ has the $r - th$ largest value $\vec{c}_{Nf}$, if $\text{sgn}(\vec{w}_f)$ is positive otherwise it has the largest negative value. It is worth noticing that the sampling idea here is motivated by [33], which however solves for a different research problem.

**Sampling Method:** The above sampling analysis leads to a simple sampling algorithm for negative context words, which is detailed in Algorithm 1.

(1) Draw a rank $r$ from a Geometric distribution.
(2) Draw a dimension $f$ from $p(f|w)$ (Eq. 14).
(3) Sort context words in terms of $c_{\cdot,f}$ in a descending order
(4) Return the word $c_N$ on position $r$ in the sorted list when $\text{sgn}(\vec{w}_f)$ is positive otherwise $c_N$ is the one ranked at $N - r$

Steps 1 and 4 has $O(1)$ complexity, step 2 includes the computation of $p(f|c)$ is $O(d)$. The only computational intensive step is 3, where items of each factor are sorted in $O(|V|log|V|)$ (i.e., $d|V|log|V|$ in total). Considering that each SGD update has little change on the overall word ranks, it is not necessary to perform step 3 for each SGD update. Empirically, we observe that recomputing the ranks every $|V| \log |V|$ iterations also yields good results. Hence, the average time complexity of step 3 is $O(d|V|log|V|)/|V|log|V|{=}O(d)$. Hence, the sampling algorithm has an amortized runtime of $O(k)$ for selecting an negative word, which is the same cost in a single gradient step of a Matrix Factorization model.

# 5 EXPERIMENTS

## 5.1 Experimental Setup

We evaluate the performance of the proposed adaptive sampler by using two real-world corpora. The first one is NewsIR[4] that is a collection of news articles derived from major newswires, such as Reuters, in addition to local news sources and blogs. The second one is the full Wikipedia articles[5]. Notice that these two training datasets are of varying sizes. The NewsIR dataset contains 30 million words. The Wikipedia 2017 dataset is about 2.3 billion words.

*Parameter Setting* We tokenize and lowercase each corpus with the Weka tokenizer. Similar to [23], the down-sampled rate is set as $1e^{-3}$, and the learning rate is set with the starting value $\eta = 0.025$ and $\eta_t = \eta(1 - t/T)$ for all experiments, where $T$ is the total number of training samples and $t$ is the number of trained samples. On both datasets, we train the skip-gram models with different samplers until it is converged.

For popularity-based sampler, we find that $power = 0.75$ offers the best accuracy. For comparison purpose, we set $window\ size = 8$, $dimension = 200$ for all methods, which are the default setting recommended in [23].

## 5.2 Evaluation Method

To begin with, we conduct experiments on two common tasks namely, word analogy and word similarity. The word analogy task is comprised of questions such as, “a is to b as c is to _?" The testing set has 19,544 such questions which are fallen into a semantic category and a syntactic category. The semantic questions are usually analogies about people name or locations. For instance, “London is to UK as Paris to _?". The syntactic questions are generally about verb tense or forms of adjectives, for example “Swim is to swimming as run is to _?". To resolve the question, the model has to uniquely capture the missing token, which means there is only one exact match that is considered as ground truth.

As for word similarity task, we use a word similarity benchmarks [7] to evaluate the correctness of our adaptive sampler. Specifically, we use the datasets collected by Faruqui and Dyer which include 7 datasets namely, SIMLEX-999, RW, WS353, MURK, WS353S, WS353R, RG65[6]. We calculate cosine value to compute the similarities between words, and then rank the similar words. The Spearman’s rank correlation coefficient is adopted to measure the correlation of ranks between human annotation and computed similarities.

To demonstrate the effectiveness of our adaptive sampler, we compare it with the original popularity-based sampling method, i.e.

[4] http://research.signalmedia.co/newsir16/signal-dataset.html
[5] https://dumps.wikimedia.org/enwiki/latest/enwiki-latest-pages-articles.xml.bz2
[6] http://www.wordvectors.org/

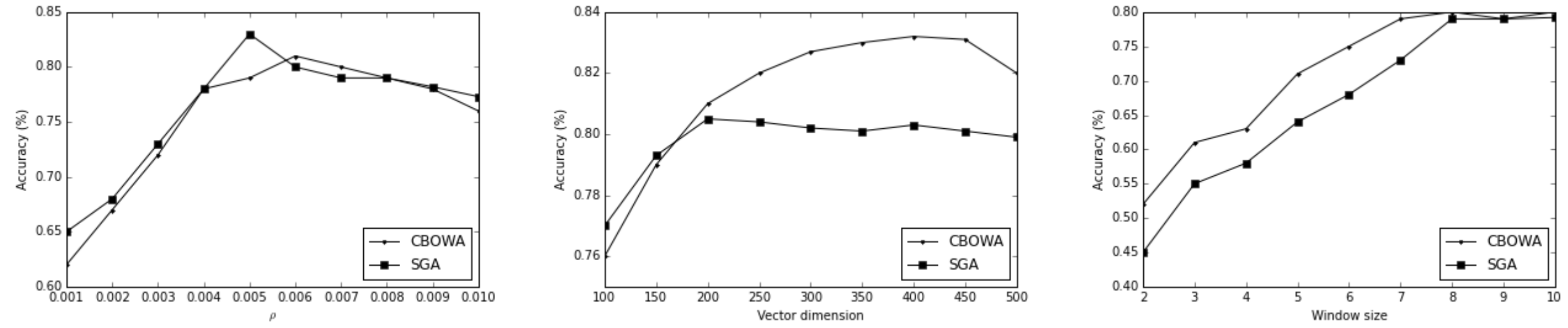

(a) (b) (c)

**Figure 3: Parameters of CBOWA and SGA of varying parameter values trained on Wiki2017 dataset with adaptive samplers**

**Table 1: The accuracy over different datasets, where $d = 200, win = 8$, and $threads = 20, neg = 5$ for both datasets (statistical significance using t-test: $^{**}$ indicates $p$-value $< 0.01$ while $^{*}$ indicates $p$-value $< 0.05$).**

| Data | **CBOW** | **CBOWU** | **CBOWA** | **SG** | **SGU** | **SGA** |
|---|---|---|---|---|---|---|
| NewsIR (1B) | 0.621 | 0.485 | 0.644 | 0.605 | 0.476 | 0.619 |
| Wiki (2B) | 0.760 | 0.592 | 0.788 | 0.782 | 0.648 | 0.793* |
| ALL (3B) | 0.768 | 0.613 | 0.792 | 0.786 | 0.654 | 0.812** |

**Table 2: The experimental results (accuracy) trained with the whole training dataset (NEWSIR+WIKI2017), where $d = 200, win = 8$, and $threads = 20, neg = 2$.**

| Data | **semantic** | **syntactic** | **total** |
|---|---|---|---|
| CBOW | 0.812 | 0.703 | 0.759 |
| CBOWU | 0.639 | 0.616 | 0.628 |
| CBOWA | 0.793 | 0.721 | 0.779* |
| SG | 0.828 | 0.794 | 0.796 |
| SGU | 0.523 | 0.537 | 0.553 |
| SGA | 0.868 | 0.798 | 0.823* |

$\rho = 0.75$. To show the effects of gradient vanishing issue, we also report results with a uniform sampler, i.e. $\rho = 0$.

- SG: The skip-gram model with the popularity-based sampler [23].
- SGU: The skip-gram model with the uniform sampler.
- SGA: The skip-gram model with the adaptive sampler described in Section 4.2.
- CBOW: Continuous bag-of-words model with the popularity-based sampler [23].
- CBOWU: Continuous bag-of-words model with the uniform sampler .
- CBOWA: Continuous bag-of-words model with the adaptive sampler.

## 5.3 Experimental Results

In order to make a fair comparison, the parameter $\rho$ of SGA and CBOWA need to be properly tuned first. The performance of word embeddings were tuned on the training set (Wiki2017) and evaluated on the testing set. The results reported in Figure 3 are those on the testing set. Figure 3 (a), (b), (c) show how the performance of adaptive sampler varies given different parameter values. From Figure 3 (a), (b), one can observe that the models achieve a good result when window size is bigger than 7 and the vector dimension is larger than 200. As mentioned in Section 4.2, $\rho$ controls the relative density of sampling distribution. From Figure 3 (c), one can see that the best performance is achieved when $\rho = 0.005$ and $\rho = 0.006$ for CBOWA and SGA, respectively.

Furthermore, [16] found that even using a small number of negative samples (e.g. $k = 5$) could achieve a respectable accuracy on large-scale datasets, although using a larger number of samples (e.g. $k = 15$) achieves considerably better performance. In Figure 5 we plot the results by increasing the number $k$ on both datasets, which shows the similar trends with [16]. As can also be seen, the accuracy on (a) converged when $k$ is larger than 15 for NewsIR dataset. One possible reason is that NewsIR dataset is noisier than Wiki2017 dataset, thereby as the number of negative pairs is increased beyond the minimum, overfitting tends to set in. We also observe that the SGA significantly outperforms SG irrespective of the number of negative pairs. Similarly, the CBOWA significantly outperforms CBOW (as shown in Fig 5) (c) and (d)), which indicates that our proposed sampling approach is more effective than the original word2vec [23].

Given the optimal parameter settings, the word analogies and word similarities tasks on the testing sets are reported in Table 1, 2 and Figure 4, respectively. First, it can be seen that our proposed adaptive sampler outperforms the classical popularity-based sampler for both the word analogies and word similarities tasks. Second, SGU has much worse prediction quality than SG and SGA.

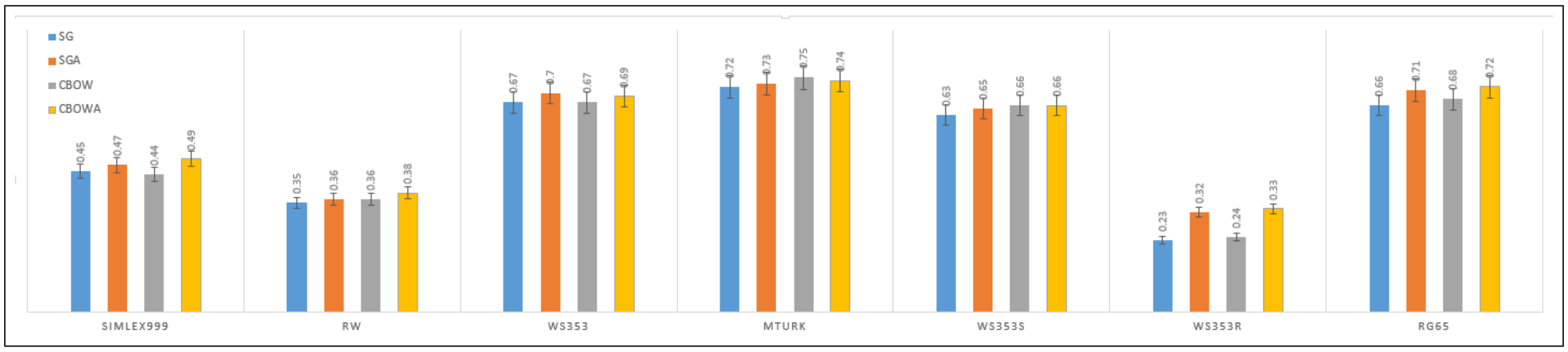

**Figure 4: Comparison of word embeddings trained with Wiki2017 for word similarity tasks on benchmark datasets**

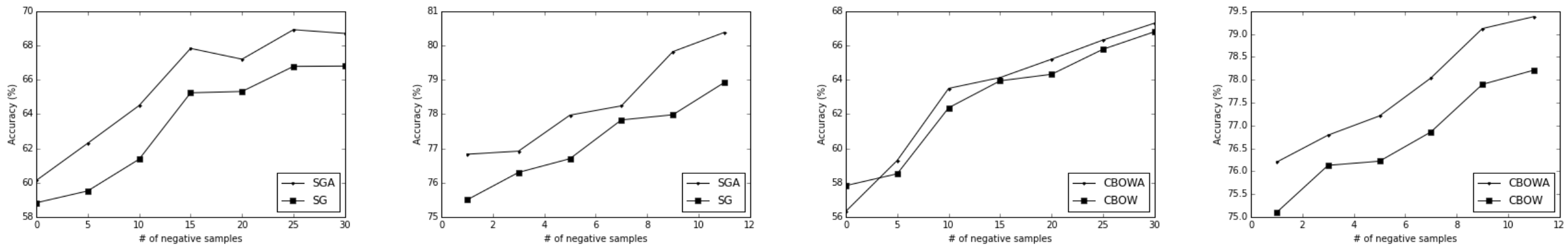

**Figure 5: (a) and (b) show the accuracy of SG with varying number of negative samples on NEWSIR and WIKI2017 datasets, respectively for the word analogy task; (c) and (d) show the accuracy of CBOW with varying number of negative samples on NEWSIR and WIKI2017 datasets for the word analogy task.**

This further verifies the gradient vanishing issue in a uniform sampler as most SGD updates have no effect on parameter changing. It also confirms that adaptive oversampling and popularity-based sampling can effectively alleviate the vanishing gradient issue.

*5.3.1 Document Classification.* As a further demonstration of the utilities of our model, we experimented with document classification with a similar setup in [27]: we use 20 Newsgroups[7] as testing set, which is a collection of newsgroup documents, partitioned evenly across 20 different newsgroups. We use the full dataset with 20 categories, such as *atheism, computer graphics*, and *computer windows X*.

*5.3.2 Runtime.* Table 3 compares training time of different samplers. All experiments are conducted on a dual 3.5GHz Intel i5-4690 machine in a single thread. The training time depends on many factors, including embedding dimension, window size, vocabulary size, and corpus size. Due to limited space, we only report the execution time with varying embedding dimension by keeping other hyper-parameters fixed. As can be seen, our adaptive oversampling does not increases the training time much. This confirms our analysis in Section 4.2 that the sampling algorithm has an amortized runtime of $O(d)$, which is the same as the costs for a single gradient step of an inner product operation.

**Table 4: The running time (minutes) of different sampling methods in the NewsIR dataset, where $win = 8$, and $threads = 20$, $neg = 25$.**

| dimension $d$ | CBOW | CBOWA | SG | SGA |
|---|---|---|---|---|
| 200 | 37.4 | 68.6 | 269.7 | 388.9 |
| 250 | 41.3 | 72.6 | 286.2 | 399.3 |
| 300 | 47.5 | 81.3 | 347.5 | 465.6 |

## 6 CONCLUSION AND FUTURE WORK

In this paper, we first elaborated the motivation of the word popularity based oversampling in word2vec [23] from both gradient vanishing and ranking perspectives. After this, we proposed an improved negative sampler that could dynamically oversample high score negative words by leveraging embedding features. The proposed *Adaptive Sampler* superseded the existing one since the sampling process took account of multi-dimensional word information instead of only popularity. More importantly, the algorithm had an amortized constant runtime and the empirical overhead is only marginal. This makes our method highly attractive for practical use.

There are several interesting and promising directions in which this work could be extended. First, in this work we only focused on two types of applications, namely, word analogy and document classification, it will be interesting to study the performance of *Adaptive Sampler* with additional tasks, such as information retrieval and question answering. it would be also interesting to investigate the performance of our sampler by applying it to complex embedding models, such as tensor factorization [39]. Finally, most existing word embedding models rely on the negative sampling techniques with an SGD optimizer, we would like to investigate more advanced optimization techniques that could handle the entire negative samples for training embedding models, e.g. in [45].

[7] https://qwone.com/ jason/20Newsgroups/

# 7 ACKNOWLEDGEMENTS

We acknowledge support from the EPSRC funded project named **A Situation Aware Information Infrastructure Project** (EP/L026015). This work was also partly supported by NSF grant #61572223.